\documentclass[preprint,12pt]{elsarticle}

\usepackage{amssymb}
\usepackage{amsthm}

\usepackage{lineno}

\usepackage[colorlinks=true]{hyperref}  %% 链接转蓝色
\usepackage{caption}
\usepackage{bbm}

\usepackage{multirow}
\usepackage{bm}
\usepackage[normalem]{ulem}
\useunder{\uline}{\ul}{}
\usepackage{subfigure}
\usepackage{algorithm}
\usepackage[noend]{algpseudocode}
\begin{document}

\begin{frontmatter}

%% Title, authors and addresses

%% use the tnoteref command within \title for footnotes;
%% use the tnotetext command for theassociated footnote;
%% use the fnref command within \author or \address for footnotes;
%% use the fntext command for theassociated footnote;
%% use the corref command within \author for corresponding author footnotes;
%% use the cortext command for theassociated footnote;
%% use the ead command for the email address,
%% and the form \ead[url] for the home page:
%% \title{Title\tnoteref{label1}}
%% \tnotetext[label1]{}
%% \author{Name\corref{cor1}\fnref{label2}}
%% \ead{email address}
%% \ead[url]{home page}
%% \fntext[label2]{}
%% \cortext[cor1]{}
%% \affiliation{organization={},
%%             addressline={},
%%             city={},
%%             postcode={},
%%             state={},
%%             country={}}
%% \fntext[label3]{}

\title{Large language model empowered participatory urban planning}

%% use optional labels to link authors explicitly to addresses:
%% \author[label1,label2]{}
%% \affiliation[label1]{organization={},
%%             addressline={},
%%             city={},
%%             postcode={},
%%             state={},
%%             country={}}
%%
%% \affiliation[label2]{organization={},
%%             addressline={},
%%             city={},
%%             postcode={},
%%             state={},
%%             country={}}

\author[label1]{Zhilun Zhou}
\author[label1]{Yuming Lin\corref{cor1}}
\ead{linyuming9@mail.tsinghua.edu.cn}
% \author[label2]{Tian Li}

\author[label1]{Yong Li\corref{cor1}}
\ead{liyong07@tsinghua.edu.cn}

\affiliation[label1]{
            organization={Department of Electronic Engineering, Tsinghua University},
            city={Beijing},
            postcode={100084}, 
            country={China}}
            
% \affiliation[label2]{
%             organization={School of Architecture, Tsinghua University},
%             city={Beijing},
%             postcode={100084}, 
%             country={China}}
            
\cortext[cor1]{Corresponding author.}

\begin{abstract}
%% Text of abstract
Participatory urban planning is the mainstream of modern urban planning and involves the active engagement of different stakeholders. However, the traditional participatory paradigm encounters challenges in time and manpower, while the generative planning tools fail to provide adjustable and inclusive solutions. This research introduces an innovative urban planning approach integrating Large Language Models (LLMs) within the participatory process. The framework, based on the crafted LLM agent, consists of role-play, collaborative generation, and feedback iteration, solving a community-level land-use task catering to 1000 distinct interests. Empirical experiments in diverse urban communities exhibit LLM's adaptability and effectiveness across varied planning scenarios. The results were evaluated on four metrics, surpassing human experts in satisfaction and inclusion, and rivaling state-of-the-art reinforcement learning methods in service and ecology. Further analysis shows the advantage of LLM agents in providing adjustable and inclusive solutions with natural language reasoning and strong scalability. While implementing the recent advancements in emulating human behavior for planning, this work envisions both planners and citizens benefiting from low-cost, efficient LLM agents, which is crucial for enhancing participation and realizing participatory urban planning.
\end{abstract}

%%Graphical abstract  %% 暂时留空，最后处理
% \begin{graphicalabstract}
%\includegraphics{grabs}
% \end{graphicalabstract}

%%Research highlights  %% 暂时留空，最后处理
% \begin{highlights}
% \item Research highlight 1
% \item Research highlight 2
% \end{highlights}


\end{frontmatter}

% \linenumbers

%% main text
\section{Introduction}

% 贡献：仿真环境下用LLM进行参与式规划（PUP）
% 传统PUP需要经验、情景、case by case，消耗时间人力物力
% RL等生成式工作尽管能快速生成，但是无法体现居民诉求，不贴近实际的工作流
% LLM能扮演个体，体现人的利益诉求，因此能作为各个利益相关方的代表，在虚拟场景下讨论
% 规划师在实际规划开始之前，可以通过LLM模拟PUP的全过程，提供规划方案供参考

Participatory Urban Planning (PUP) is a process of collaborative decision-making that involves the active engagement of different stakeholders to tackle complex spatial challenges and achieve sustainable urban development~\cite{arnstein1969ladder, forester1982planning, forester1999deliberative}. This inclusive approach aims to incorporate diverse perspectives and promote a sense of ownership among community members in planning and decision-making processes, making it one of the most commonly used methods in current urban community planning~\cite{li2020collaborative}.

The traditional participatory urban planning method requires extensive experience, consideration of multiple interests, and often a case-by-case examination of unique community problems. Therefore, it demands significant time and manpower and heavily relies on experienced urban planners~\cite{eriksson2022opening}. As cities continue to grow and the need for urban renewal steadily increases, the limitations of traditional urban planning methods become more apparent, prompting the exploration of innovative methodologies that can enhance their efficiency and effectiveness~\cite{tian2023participatory}.

Various generative urban planning techniques have been proposed in the literature, such as evolutionary algorithms~\cite{koenig2020inteegrating}, Generative Adversarial Networks (GAN)~\cite{quan2022urban}, Variational Autoencoder (VAE)~\cite{wang2021deep}, or Reinforcement Learning (RL)~\cite{zheng2023spatial, qian2023ai}. While these methods have shown promise in generating planning solutions, they usually take objective optimization goals and lose sight of the nuanced demands of residents, which may harm the interests of vulnerable groups and is usually hard to adjust. The inherent limitations of these generative methods highlight the need for an approach that encapsulates human-centric considerations.

Large Language Models (LLMs) have emerged as a promising solution to these challenges. These advanced natural language processing models, such as GPT-3, are capable of understanding and generating human-like text for various tasks~\cite{brown2020language}. They are trained on colossal datasets, which makes them highly versatile in language understanding abilities and is a useful tool for content creation. Therefore, it is possible to generate an urban planning scheme using natural language interaction with the LLM, fulfilling the human-centric considerations easily. Another significant advantage of LLMs is that they can create human-like agents, which can mirror the interests and demands of people with simple role-play prompts~\cite{shanahan2023role}. Based on the role settings, LLM agents can carry out complex reasoning and interactions~\cite{hagendorff2023human}, and even run a virtual town with events like birth party and mayor election~\cite{park2023generative}. This unique ability makes them effective representatives of diverse stakeholders in virtual scenario discussions and contributes to a comprehensive and inclusive decision-making process in a virtual way.

Integrating LLM into participatory urban planning processes introduces a new approach that allows the governor and planners to simulate the entire process before actual planning activities begin. Besides the ability to generate reasonable planning based on their wealth of knowledge, LLM agents can serve as virtual representatives of diverse stakeholders, pre-planning simulations provide a mechanism to explore potential public engagement and resident feedback, providing valuable insights and references. The possible planning solutions can also be generated in a discussion loop with proper design. The ability to anticipate and address potential needs and challenges in a simulated environment can save significant time and manpower, highlight the interest of under-representative vulnerable populations, and enhance the robustness of subsequent planning stages before the participatory process starts.

In conclusion, our study introduces a novel urban planning approach leveraging the capabilities of LLM within the participatory urban planning paradigm. The findings demonstrate that LLM can generate coherent urban planning schemes when provided with appropriate prompts. Moreover, integrating LLM into the participatory planning workflow, encompassing role-play, collaborative generation, and feedback iteration, enhances inclusivity and decision-making efficiency. Empirical experiments in two communities reveal that LLM surpasses human experts in need-agnostic service and ecology metrics, and achieves optimal results in need-aware satisfaction and inclusion metrics. Detailed analysis underscores that the LLM model effectively addresses the limitations inherent in generative methods by considering individual needs and delivering transparent, easily adaptable outcomes. Subsequent sections will delve into discussions concerning related works, method design, experimental settings, and the evidenced results that substantiate the efficacy of our framework.

\section{Related work}

% 已有的参与式规划
% 生成式规划（CNN、图、RL）
% LLM已有的成果

\subsection{Participatory planning}

Participatory Planning has evolved as a pivotal approach in urban planning, emphasizing community engagement in the decision-making process~\cite{forester1999deliberative}. Tracing the evolution of participatory planning from its early roots, seminal works by Arnstein~\cite{arnstein1969ladder} and Forester~\cite{forester1982planning} underscore the fundamental principles of citizen involvement in planning processes. 

Although scholars emphasize the imperative of deliberative democracy and the empowerment of marginalized voices, the practical application of these theories encounters formidable challenges in translating principles into actionable practices. An empirical study by Monno and Khakee~\cite{monno2012tokensim} indicates that planners often lean towards restricting participation to mere information and consultation, lacking assurance that concerns will be genuinely considered. Professionals may express nominal support in the abstract but harbor reservations about citizens' qualifications, citing a perceived lack of commitment, competence, or foresight in their interests~\cite{astrom2020participatory}. In the context of rural Indonesia, Akbar et al.~\cite{akbar2020participatory} illustrate a procedure heavily reliant on local elites, resulting in limited representativeness and the exclusion of marginalized groups, including the disabled and impoverished. As a summary, Abas et al.~\cite{abas2023systematic} point out the five main challenges for public participation, including cost, lack of skilled facilitators, low efficiency, low interest to participate, and language barriers. 

Fortunately, interactive dashboards, Augmented Reality (AR), Virtual Reality (VR), and Artificial Intelligence (AI) can help the practice of participatory democracy. Online participatory tools featuring data and figures give participants an accessible platform to articulate their views and engage in real-time discussions, thereby alleviating impediments to information dissemination~\cite{afzalan2018online, lock2020review}. Moreover, lots of digital participatory planning practices show AR and VR can offer immersive and inclusive environments, facilitating diverse stakeholder participation and effectively addressing challenges related to travel costs, language barriers, and professional thresholds~\cite{jiang2018demonstrator, sasmannshausen2021citizen, chassin2022experiencing, nasrazadani2022rapid,  ahmadi2023augmented}. Du et al.~\cite{du2023artificial} contend that AI can be helpful in participant selection, identifying domain experts, and weighing opinions, thereby easing the burden on facilitators and promoting efficiency. Additionally, AI chatbots, as demonstrated in an experiment in Afghanistan~\cite{haqbeen2021using}, prove effective in encouraging people to share opinions, achieving response rates comparable to human facilitators. 

However, despite the promising potential of technological innovations to enhance citizen engagement, there remains a significant gap in truly transforming group processes and outcomes~\cite{du2023artificial}. The limited familiarity and a dearth of consistency and transparency restrain the more substantial integration of AI methods~\cite{lock2021towards}. The ongoing progress in natural language processing holds the promise of addressing these challenges and unlocking the full potential of technological interventions in participatory processes.

\subsection{Multi-agent collaboration with large language models}

There has been an adequate study showing that LLM can serve as an emulated role-play tool~\cite{shanahan2023role}. Many studies accompanying the recent advancement of LLM now focus on multi-agent collaboration with LLMs, where different roles will be assigned to LLM agents. Then, these agents will cooperate to solve a complex task together. 

Specifically, LLM agents are usually assigned roles based on distinct expert knowledge in the role-play part. For example, solving operations research problems needs roles like terminology interpreter, modeling expert, programmer, and evaluator~\cite{anonymous2024chainofexperts}, while developing software may need product manager, architect, project manager, engineer, and QA engineer~\cite{hong2023metagpt}. Moreover, some studies craft LLM agents with adaptive roles when solving the problem~\cite{chen2023agentverse}. To enable LLM agents to act in specific roles, researchers usually carefully design prompts to guide agents' behavior and provide expert knowledge needed by prompting or external knowledge bases.

After the role-play, collaboration mechanisms are designed to enable the cooperation of different agents. Such a mechanism can be the form of a pipeline, where different agents sequentially finish part of the task~\cite{hong2023metagpt,anonymous2024chainofexperts,yu2023thought}, or group discussion, where several agents communicate with each other to reach an agreement~\cite{wu2023autogen,zhang2023exploring}, or the hybrid of them~\cite{chen2023agentverse}. 

Multi-agent collaboration with LLMs has shown considerable success in many domains, including operations research~\cite{anonymous2024chainofexperts}, text quality evaluation~\cite{chan2023chateval}, solving math problems~\cite{li2023camel}, and software development~\cite{hong2023metagpt}. However, none of the existing studies have applied LLMs to simulate citizens in urban planning scenarios. Moreover, most existing works only incorporate a few LLM agents, which cannot be directly applied to participatory planning where a community may contain a dozen thousand residents.

\section{Methods}

% 图1：framework
% LLM调用相关技术
% 角色扮演：规划师、居民画像
% 讨论机制：分层分区讨论、模仿PP的讨论技巧
% 反馈迭代

\subsection{LLM and the overall framework}

As indicated by prior research, LLMs exhibit remarkable natural language comprehension across diverse domains. Therefore, a straightforward approach involves framing planning challenges in natural language and using the multi-modal capabilities of the large language model to interpret planning instructions and generate design. Besides, the expansive scope of natural language makes LLM valuable participants as they can assume a quasi-human agent role. By crafting prompt templates aligning with specific roles, LLM agents can emulate professional planners or local residents, obviating the need for intricate manual design of internal mechanisms. While the exact mechanisms underpinning LLMs' capabilities remain partially understood, their extensive textual and real-world knowledge empowers them to make contextually rational decisions, rendering them efficacious participants in participatory processes.

\begin{figure}[htbp!]
    \centering
    \includegraphics[width=.9\linewidth]{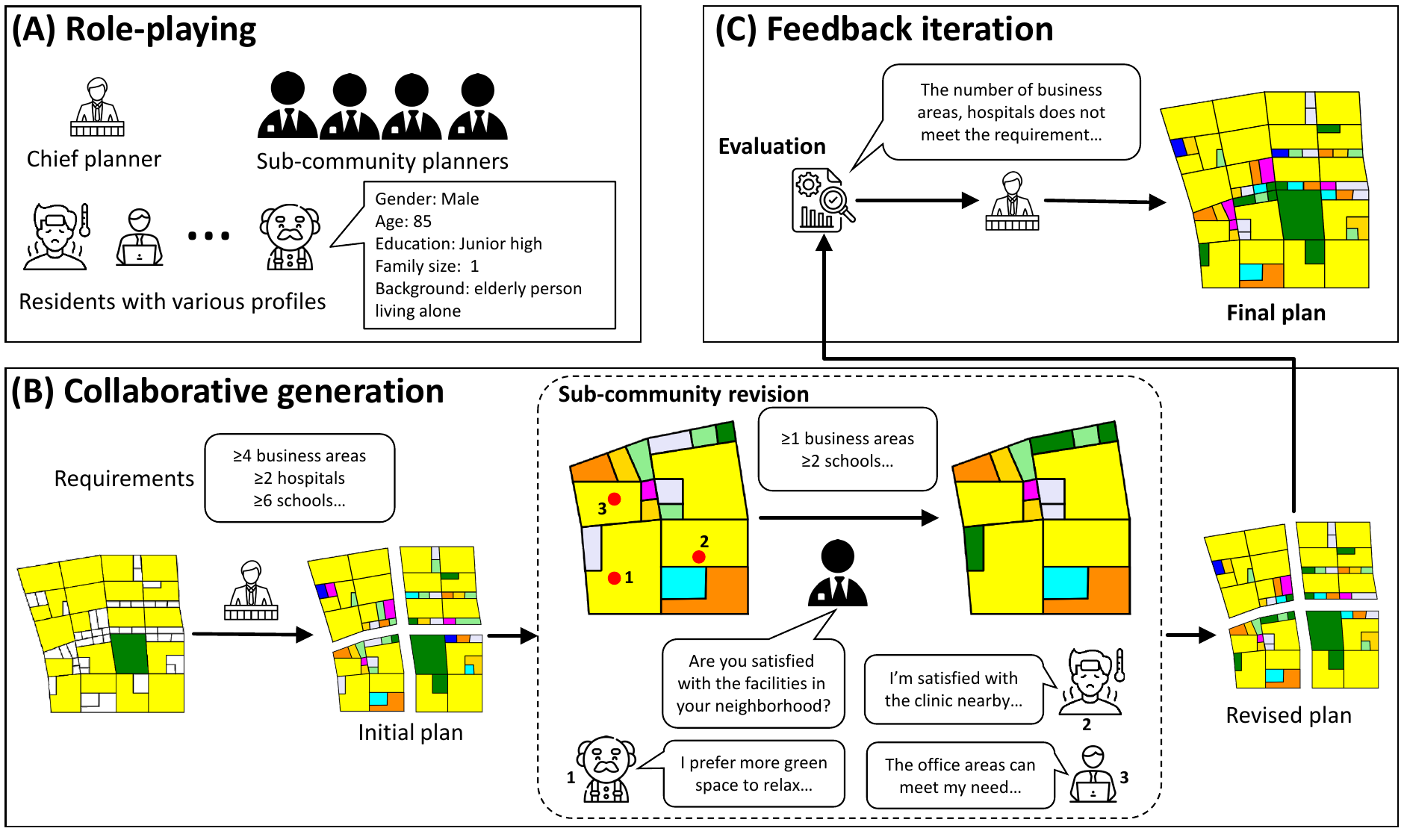}
    \caption{The framework of our proposed participatory planning method.}
    \label{fig:framework}
\end{figure}

To systematically reveal the latent capabilities of LLM agents, we have formulated a participatory planning framework harnessing the generative prowess of LLM, as depicted in Figure~\ref{fig:framework}. This framework adheres to conventional participatory planning, incorporating simplifications for enhanced representational clarity. It encompasses three primary modules: role-playing, collaborative generation, and feedback iteration.

\subsection{Role-playing}

In leveraging the role-playing capability of LLM, we craft diverse agents to emulate distinct real participants, equipping them with fundamental attributes and pertinent knowledge (Figure~\ref{fig:framework}A). Adopting the prevalent organizational structure of planning teams, we designate a chief planner (CP) responsible for overarching planning considerations, alongside several sub-community planners (SP) tasked with in-depth exploration of regional challenges. The transformation of all planning actions into natural language enables both CP and SP to formulate a planning scheme independently. For the residents, we emulate their foundational attributes to generate reliable agents aligning with their respective profiles.

It is essential to highlight the adaptability and flexibility of all agents, allowing for seamless adjustment and extension of functionalities through straightforward natural language commands. For conciseness, the present framework excludes other stakeholders like government officials or estate developers; nevertheless, their integration can be accommodated effortlessly. Additionally, the interaction with agents occurs exclusively in natural language, facilitating the substitution of these agents with real individuals and ensuring a smooth transition between the simulated environment and real-world scenarios.

\subsection{Collaborative generation}

Building upon these LLM agents, we initiate collaborative planning generation, using community-level land-use redevelopment as an illustrative case for generality (Figure~\ref{fig:framework}B). The collaborative generation process involves distinct key steps. Initially, the CP agent, leveraging its intrinsic professional knowledge and multimodal capabilities, crafts a proposal for community land use with the allocation for each plot. Subsequently, this proposal undergoes iterative discussions facilitated within the participatory planning framework, engaging resident agents with diverse attributes and vested interests. Then, the discussion results are relayed to the CP agent for a new proposal.

Note that the discussions for each plot are moderated by an SP agent, who solicits opinions and opposition from resident agents in the respective areas. These discussions encompass not only the rebuttals of existing proposals but also involve residents engaging in mutual discussions, disputes, and persuasion based on divergent interests. Despite the SP agent overseeing the entire discussion, their primary role is summarizing and organizing the ultimate result derived from the residents. The outcome is subsequently fed back to the CP agent, guiding corresponding revisions to ensure the substantive inclusion of resident perspectives. Therefore, our framework transcends tokenism consultative participation, embodying substantive content controlled by the citizen members.

\subsection{Feedback iteration}

The feedback iteration component, illustrated in Figure~\ref{fig:framework}C, is designed to address the deficiencies in the planning scheme, exemplified by the predetermined number of land-sue. Following the revision of the proposal based on resident perspectives, unintended violations of crucial planning directives may occur, such as insufficient allocation of hospitals. These violations are conveyed to the CP through natural language templates, ensuring attention to these constraints in subsequent solution discussions. This approach is intuitive, considering the planner must balance planning constraints with the residents' demands.

\subsection{Output}

The primary and most tangible outcome of our model is participatory-generated land-use planning. The final plan represents a novel generative approach to urban planning, offering valuable references for practical planning initiatives. A notable advantage of this generative model is its capacity to swiftly produce numerous plans autonomously, and the underlying rationale of the generated plan can be carefully examined, which is hard for other generative methods.

Additionally, the LLM agents engendered in this process are also noteworthy outputs. Operating on natural language and adaptable in practical applications, both planner and resident agents can be seamlessly integrated into existing participatory planning processes. Planners can experiment by proposing planning solutions and assessing the sentiments and objections of resident agents, facilitating efficient collection of essential feedback before the onset of the participatory process. Similarly, residents keen on participatory planning can initiate communication with planner agents to efficiently grasp planning knowledge, enhancing their ability to articulate interests and concerns effectively within the participatory planning process.

% The APIs are requested asynchronously so that multiple APIs can be called and their responses received concurrently. The models called via the toolkit include all the models in our work that are accessible by API, namely those that are commercial rather than open-source: gpt-4, gpt-3.5, bard, claude-instant, text-bison, text-davinci-002, and text-davinci-003. 
% The models that can accept image inputs are bard and gpt-4, whereas the rest accept exclusively text. 
% The API interface that is requested for each model is in some cases is made available by the its provider, thereby allowing developers to integrate theirs into a versatile array of applications easily. However, in other cases, third-party APIs are used. Specifically, gpt-4, gpt-3.5, text-davinci-003, and text-davinci-003 are accessed via OpenAI's platform\footnote{https://openai.com}, text-bison is accessed via Google's PaLM API interface\footnote{https://ai.google.dev/models/palm}, bard is queried via a third-party API\footnote{https://github.com/dsdanielpark/Bard-API}, and finally, claude-instant is accessed via Poe\footnote{https://poe.com}. 

\section{Experiment}

% 数据集：回龙观、大红门。（图2：初始地图、卫星、待规划；表1：统计信息）
% 实验设置
% 评价指标

\subsection{Study area}

\begin{figure}[htbp!]
    \centering
    \includegraphics[width=.8\linewidth]{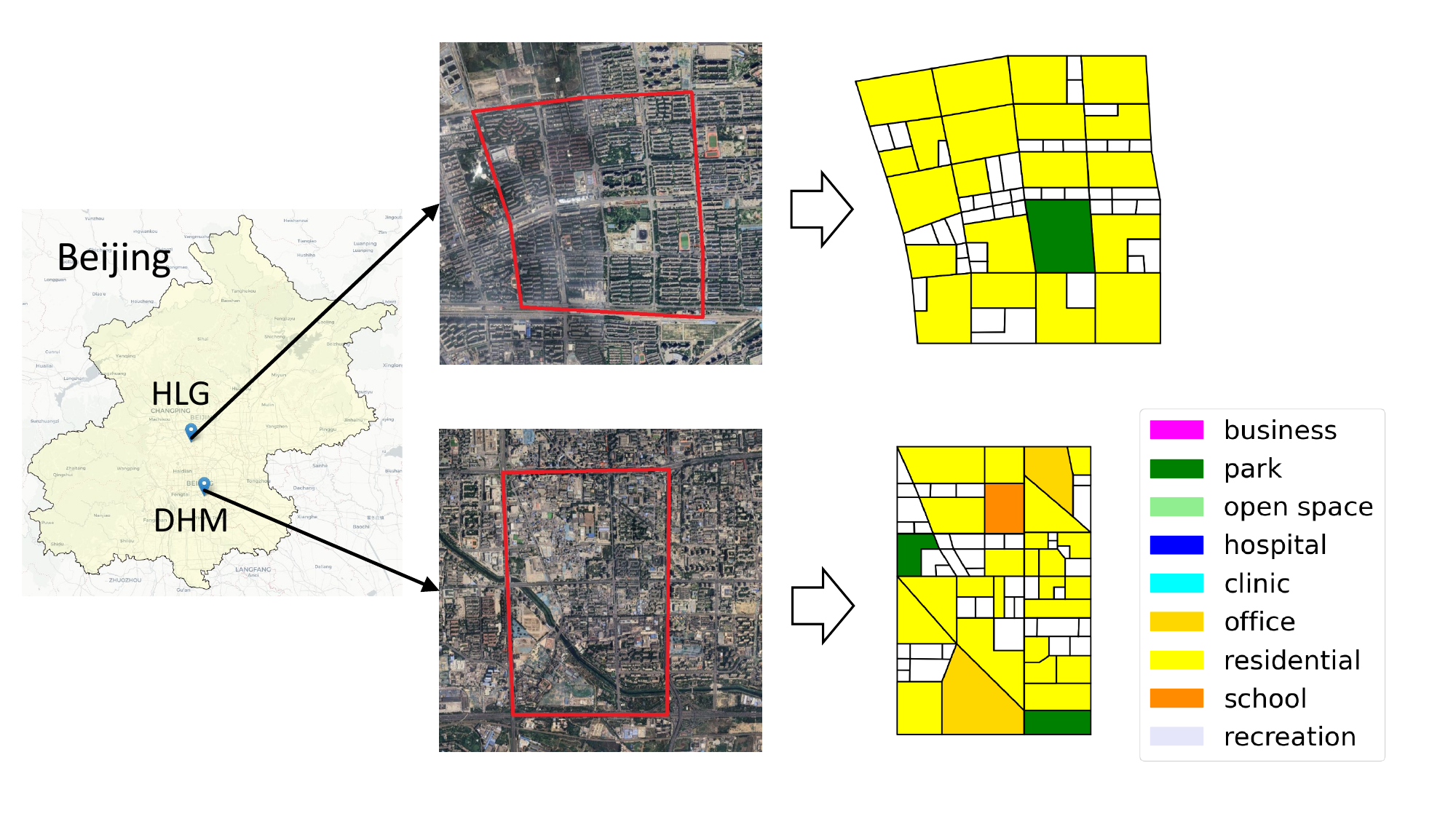}
    \caption{The satellite image and area division of two communities, Huilongguan (HLG) and Dahongmen (DHM).}
    \label{fig:dataset}
\end{figure}

The experimental study was carried out in two representative communities, Huilongguan (HLG) and Dahongmen (DHM), which serve as distinct representations of urban models in Beijing (see Fig.~\ref{fig:dataset}). HLG, situated 33km north of Beijing's city center, was originally a satellite town and has now integrated into the Beijing metropolis within the 6th Ring Road. Characterized by extensive high-rise residential complexes, HLG has evolved into one of the most densely populated communities in Beijing. However, the surrounding infrastructure is inadequate, and limited employment opportunities necessitate many residents to commute for remote work, making HLG a typical commuter town.

Conversely, DHM, located in south Beijing, represents a different urban landscape. The name "Dahongmen" or "Big Red Gate" originally referred to a gate in the Imperial Garden dating back to the 16th century. With a long history of residency and the Liangshui River traversing its expanse, Dahongmen was historically the largest clothing wholesale trading center in North China. Over time, the area has experienced unplanned and haphazard development, resulting in a scattered urban layout that intertwines residential spaces with commerce, warehousing, logistics, and rental compounds. Apart from the disordered landscapes, some informal settlement areas lacking sufficient infrastructure have been preserved, rendering the area susceptible to flooding and other risks.

The choice of communities considers their diverse socio-economic profiles, demographic compositions, and complexities in urban planning. Key information about these two areas, derived from the census yearbook~\cite{BMBS2022Beijing}, is presented in Table~\ref{tbl:dataset}. HLG, being a relatively recent community, demonstrates a predominantly youthful resident demographic, with a significant 48.88\% holding a Bachelor's degree or higher. In contrast, DHM has a higher proportion (24.23\%) of elderly residents aged over 60, introducing distinct preferences and motivations in participatory urban planning. By selecting communities with diverse resident profiles, our experiment aimed to capture a broad spectrum of participatory urban planning scenarios, fostering a comprehensive understanding of the LLM's effectiveness in diverse contexts.

\begin{table}[htbp!]
    \centering
    \caption{The basic information of two real-world datasets.}
    \resizebox{.95\linewidth}{!}{
        \begin{tabular}{c|cccccc}
        \hline
        \textbf{Community} & \textbf{Area} & \textbf{\#Residents} & \textbf{Age>60} & \textbf{Bachelor's degree} & \textbf{\#Plots} & \textbf{\#Vacant Plots}\\
        \hline
        HLG &   3.74 km$^2$ &   85,041 &   16.38\% &   48.88\% &   63  &   42  \\
        DHM &   5.17 km$^2$ &   73,130 &   24.23\% &   31.54\% &   70  &   42  \\
        \hline
        \end{tabular}
        }
    \label{tbl:dataset}
\end{table}

\subsection{Initial conditions and planning requirements}

We replicate the initial conditions of the HLG and DHM communities using OpenStreetMap data. To simplify the complex spatial setting, we partitioned the areas into plots based on roads and existing land-use conditions by expert input. Following common scenarios in actual urban redevelopment, we retain the residency and major green land plots and treat another land-use plot as vacant spaces. Both HLG and DHM consist of 42 vacant plots that can be re-zoned into eight different functionalities, including school, hospital, clinic, business, office, recreation, park and green space. The proposed LLM framework simulates the roles of CP, SP, and residents in the participatory diagram to redevelop the areas.

To ensure that the communities are realistic and well-planned, we established some basic requirements. For both communities, a minimum number of plots are mandated to ensure adequate infrastructure coverage: 6 plots for schools, 6 for recreation areas, 4 for businesses, 4 for clinics, and 1 for parks and green spaces. Reflecting the need for job opportunities in HLG, the minimum number of office plots is 6, while in DHM is 2. Addressing the needs of the aging population, a minimum of 2 hospital plots are required in HLG, compared to 1 in DHM.

To align with reality, the background of participants is considered. We created 1,000 resident agents for each community, balancing the actual population and simulation capabilities. The agent profiles, encompassing gender, age, family size, and education, are sampled from statistical distributions. It is noteworthy that the agent number surpasses 1\% of the actual population in both communities, aligning with the standard used in yearly censuses and exceeding the typical consideration in participatory planning. We also intentionally included four vulnerable groups by creating additional agents: families with children, families with patients, elderly individuals living alone, and rental migrants. For each community, one CP agent and four SP agents were created.

\subsection{Evaluation metric}

To evaluate the performance of our framework, we employed two categories of metrics: need-agnostic and need-aware. Need-agnostic metrics are aggregated indicators providing an overview of the entire community, focusing on service and ecology without individual need. For service, the minimum distance $d(m, j)$ for the agent $m$ to access the $j$th type of plots $\mbox{P}^j$ is calculated as:
\[
    d(m, j) = \min \{ \mbox{EucDis}(\mbox{L}_m, \mbox{P}_1^j), \dots, \mbox{EucDis}(\mbox{L}_m, \mbox{P}_{k_j}^j) \}, 
\]
where $\mbox{L}_m$ represents the home location of agent $m$, and the $k_j$ is the total number of plots of type $j$. Subsequently, the Service metric is defined as:
\[
    \mbox{Service} = \frac{1}{n_m} \sum_{m=1}^{n_m} \frac{1}{n_j} \sum_{j=1}^{n_j} \mathbbm{1}[d(m,j)<500],
\]
where $n_m$ denotes the number of agents and $n_j$ is the number of land use types. The $\mbox{Service}$ metric signifies the proportion of accessible services within a 500m radius, ranging from 0 to 1.

On the other hand, the Ecology Service Range (ESR) is defined as the union of buffers extending 300m from each park and open space $\mbox{P}_k^{park}$, represented as:
\[
    \mbox{ESR} = \mbox{Union} \{ \mbox{Buffer}(\mbox{P}_1^{park}, 300), \dots, \mbox{Buffer}(\mbox{P}_{k}^{park}, 300) \} ,
\]
subsequently, the Ecology metric is defined as:
\[
    % \mbox{Ecology} = \frac{\sum_{i}{\mbox{Area}(\mbox{R}_i \cap \mbox{ESR})}}{\sum_{i}\mbox{Area}(\mbox{R}_i)},
    \mbox{Ecology} = \frac{1}{n_m}\sum_{m=1}^{n_m}\mathbbm{1}[\mbox{L}_i \in \mbox{ESR}],
\]
where $\mbox{Ecology}$ signifies the proportion of agents covered by the ecological service range, ranging from 0 to 1. The two need-agnostic metrics effectively encapsulate the concept of a 15-minute life circle~\cite{smartcities4010006}, ensuring that basic community services are reachable within 15 minutes by walking or cycling. It's noteworthy that the flexibility of metric selection allows for further extensions in real-world scenarios.

However, the $\mbox{Service}$ and $\mbox{Ecology}$ metrics do not consider the various needs of residents with different profiles. Therefore, we further introduce two need-aware metrics $\mbox{Satisfaction}$ and $\mbox{Inclusion}$.
At the forefront of participatory planning, the paramount consideration lies in meticulously examining residents' feedback and opinions. Within our simulated environment, each resident agent $m$ can report a set of land-use types comprising 3-5 deemed most urgently needed, denoted as $J_m$. We define the satisfaction level for agent $m$ as a need-aware metric:
\[
    S_m = \frac{1}{n_j} \sum_{j=1}^{n_j} \mathbbm{1}[d(m,j)<500], \quad j \in J_m ,
\]
where $d(m,j)$ represents the minimum distance for the $m$th agent to access the $j$th type of plots. The overall satisfaction metric is then calculated as:
\[
    \mbox{Satisfaction} = \frac{1}{n_m} \sum_{m=1}^{n_m} S_m,
\]
with $n_m$ denoting the total number of agents. The $\mbox{Satisfaction}$ metric quantifies the extent to which each agent's needs are fulfilled, ranging from 0 to 1. In parallel, we introduce an Inclusion metric to safeguard the interests of marginalized groups $V$ in a similar way:
\[
    S_v = \frac{1}{n_j} \sum_{j=1}^{n_j} \mathbbm{1}[d(v,j)<500], \quad j \in J_v , \quad v \in V,
\]
\[
    \mbox{Inclusion} = \frac{1}{n_v} \sum_{v=1}^{n_v} S_v.
\]
The $\mbox{Inclusion}$ measures whether the planning process adequately addresses the requirements of marginalized groups, contributing to a more equitable and inclusive urban development strategy. We combine the need-agnostic $\mbox{Service}$ and $\mbox{Ecology}$, and need-aware $\mbox{Satisfaction}$ and $\mbox{Inclusion}$ to measure the performance of the algorithm, aiming to in line with the core aspirations of participatory planning and taking care of the satisfaction of all stakeholders.

\subsection{Baseline}

To evaluate our framework of participatory urban planning environments, we selected six methods as baselines for comparison: the random method, the centralized method, the decentralized method, the Geometric Set Cover Algorithms (GSCA), the Deep Reinforcement Learning (DRL), and the result from human designers. Specifically:

\begin{itemize}
    \item Random: Each plot will be randomly assigned a functionality ensuring the minimum requirement.
    \item Centralized: For each functionality, the probability of assigning a new plot is inversely proportional to the distance to the community center.
    \item Decentralized: For each functionality, the probability of assigning a new plot is proportional to the distance to the same type of plot.
    \item GSCA: For each functionality, solve the geometric-set-coverage-like problem by maximizing the coverage of the given facility type.
    \item Human Expert: Recruit professional human designers to accomplish the planning tasks from the same initial condition.
    \item DRL: Using the deep reinforcement learning method by Zheng et al.~\cite{zheng2023spatial} to maximize the metric \mbox{Service} and \mbox{Ecology}.
\end{itemize}

The implementation of baseline methods follows established procedures outlined in the literature~\cite{zheng2023spatial}. A team of 8 professional planners from the UK and China, with a minimum of 3 years of experience in urban planning, was recruited for the study. The results generated by human designers were transformed in ArcGIS format, enabling the subsequent computation of relevant metrics.

\section{Results}

% 主要结果： random、rule-based、人类、（RL）、ours （图3 结果及打分比较）

% R1说明roleplay有效（2页）：

% 	Roleplay影响
% 	文本分析（图3：对话图）

% R2说明讨论有效（2页）：
% 图：讨论文本
% 	规划图对比，分析修改原因（图5：讨论修改了哪些）；
% 	讨论前后指标对比（图6：折线图显示主要讨论指标变化）

% R3反馈机制（2页）：
% 	反馈过程的文本图（图7：文本）
% 	讨论过程的分析，那些规划元素在PP中被强调了，是否符合PP预期

% R4人类实验（2页）：（暂时不加）
% 	将讨论过程、规划结果呈现给规划师，认为是否对PP开展有帮助（表3）
% 	包括了解当地情况、居民困难、规划诉求、说服技巧、整体有效（参考其他规划辅助工具）

Table~\ref{tbl:result} presents the outcomes of the experiments conducted in the HLG and DHM communities, comparing them with the baseline methods. Instances where the CP Agent generated proposals without engaging in participatory planning discussions are labeled as "Ours w/o discuss." Generally, our method outperformed most rule-based and human-designed solutions in need-agnostic metrics while showing a slight gap compared to RL. Besides, our method notably surpassed in need-aware metrics.

\begin{table}[htbp!]
    \centering
    \caption{Performance comparison with baselines on two datasets. The best results are presented in bold, and the second-best results are underlined.}
    \resizebox{.95\linewidth}{!}{
\begin{tabular}{c|cccc|cccc}
\hline
                       & \multicolumn{4}{c|}{\textbf{HLG}}                                             & \multicolumn{4}{c}{\textbf{DHM}}                                              \\
\textbf{Model}         & \textbf{Service} & \textbf{Ecology} & \textbf{Satisfaction} & \textbf{Inclusion} & \textbf{Service} & \textbf{Ecology} & \textbf{Satisfaction} & \textbf{Inclusion} \\ \hline
\textbf{Random}        & 0.491            & 0.505            & 0.708           & 0.698                 & 0.690            & 0.664            & 0.691           & 0.701                 \\
\textbf{Centralized}   & 0.654            & 0.364            & 0.578           & 0.560                 & 0.562            & 0.393            & 0.526           & 0.539                 \\
\textbf{Decentralized} & 0.709            & 0.455            & 0.678           & 0.691                 & {\ul 0.743}      & 0.518            & 0.694           & 0.703                 \\
\textbf{GSCA}          & 0.682            & 0.439            & 0.653           & 0.657                 & 0.584            & 0.464            & 0.605           & 0.616                 \\
\textbf{Human Expert}  & 0.713            & 0.586            & 0.692           & 0.714                 & 0.633            & 0.723            & 0.750           & 0.755                 \\
\textbf{RL}            & \textbf{0.773}   & \textbf{0.747}   & 0.708           & 0.716                 & 0.671            & \textbf{0.880}   & 0.576           & 0.597                 \\ \hline
\textbf{Ours w/o discuss}  & 0.75             & {\ul 0.746}      & {\ul 0.735}     & {\ul 0.719}           & 0.729            & {\ul 0.755}      & {\ul 0.756}     & {\ul 0.777}           \\
\textbf{Ours}          & {\ul 0.756}      & 0.714            & \textbf{0.784}  & \textbf{0.764}        & \textbf{0.760}   & 0.739            & \textbf{0.784}  & \textbf{0.794}        \\ \hline
\end{tabular}
        }
    \label{tbl:result}
\end{table}

In the HLG experiment, our method achieved a Service metric score of 0.756, denoting that residents can access 75.6\% of necessary facilities within a 500m radius, ranking second only to RL's 0.773. On the Ecology metric, our initial proposal received a score of 0.746, indicating that 74.6\% of residential areas fall within the 300m service range of green spaces, closely approaching RL's 0.747. However, after participatory discussions, the Ecology score decreased to 0.714, but still surpassed the human expert's score of 0.586. Considering that increasing facilities may reduce green spaces, a potential trade-off between Service and Ecology may exist. Similar observations were made in DHM, where our approach achieved the top score of 0.760 on the Service metric after discussions with residents, while the Ecology score slightly decreased from 0.755 to 0.739, not reaching RL's 0.880. Despite the potential trade-off, our approach demonstrated need-agnostic metrics second only to RL, surpassing human designers.

Conversely, our approach unequivocally achieved the first position in need-aware metrics. In HLG, our final solution obtained a Satisfaction score of 0.784, signifying that 78.4\% of residents' diverse needs were satisfied. The Inclusion score was 0.764, indicating that 76.4\% of vulnerable groups' needs were satisfied, remaining the highest among all solutions. In DHM, our solution similarly excelled with scores of 0.784 and 0.794, indicating optimal fulfillment of each resident agent's specific needs. Notably, RL, which performed well on need-agnostic metrics, did not excel in need-aware metrics, falling short of the results by human planners based on experience. Overall, discussions with resident agents significantly enhanced performance in Satisfaction and Inclusion, substantiating the substantive incorporation of residents' opinions across experiments.

\subsection{Analysis of the role-play}

\begin{figure}[htbp!]
    \centering
    \includegraphics[width=.9\linewidth]{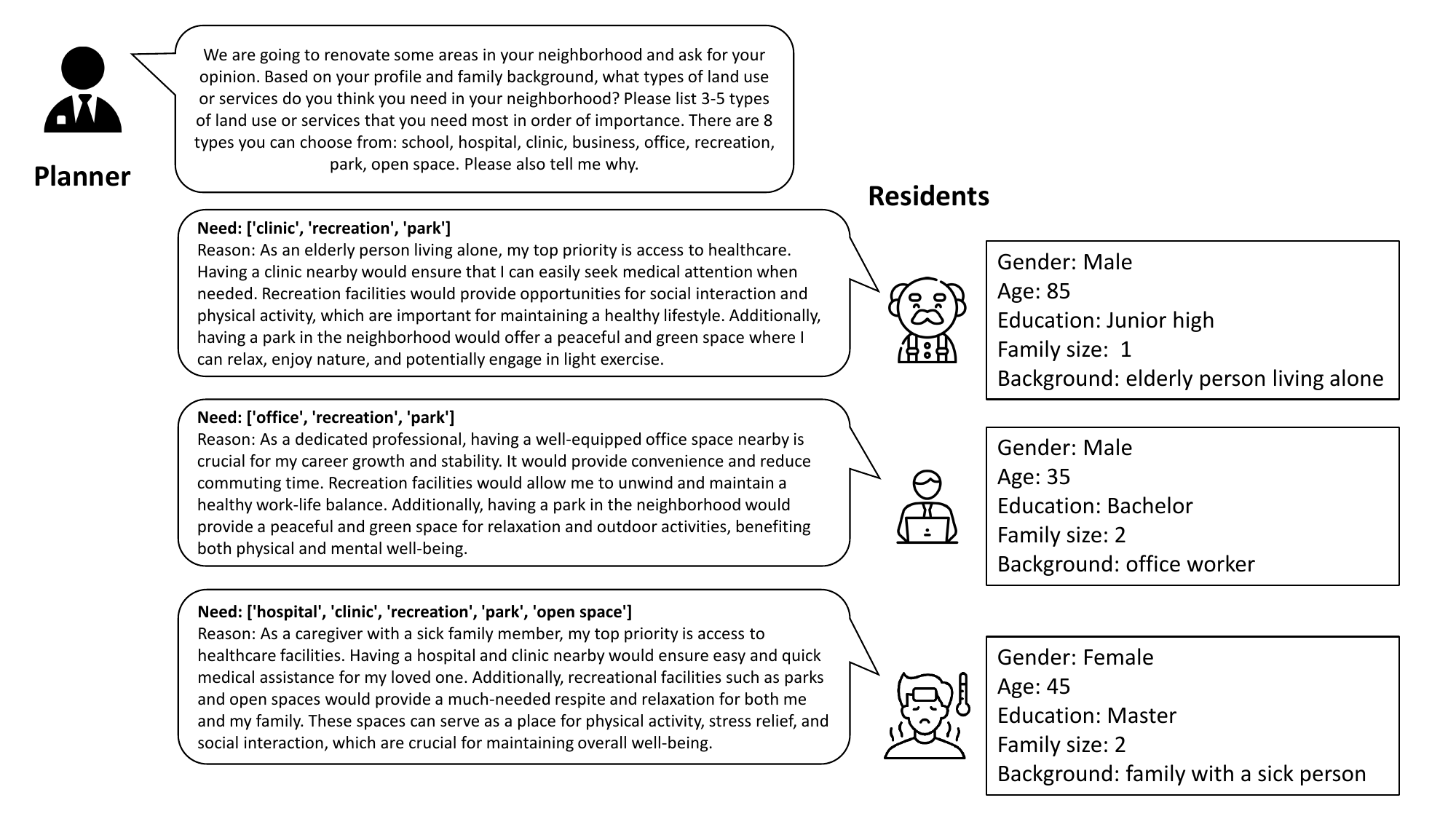}
    \caption{Dialogue of asking residents' needs of facilities. Different residents have different needs corresponding to their profiles and special backgrounds.}
    \label{fig:ask_need}
\end{figure}

As LLM operates on natural language, a comprehensive examination of the inputs and outputs of each module allows us to verify their alignment with the intended functionality. Figure~\ref{fig:ask_need} illustrates examples where each resident agent generates distinct land-use planning demands based on their profiles, along with easily understandable justifications. As expected, the elderly living alone may prioritize their physical health, seeking proximity to the clinic. Workers may prefer nearby office spaces, and patients may desire proximity to hospitals and clinics. While these considerations might appear to be common sense, it is crucial to highlight that addressing such specific demands has been challenging for previous methods. Acquiring these highly individualized requirements typically involves extensive interviews, and the process at the scale of 1000 is impractical since surpasses the capacity of human planners. Our results demonstrate that LLM agents can effectively comprehend human profiles and appropriately role-play based on their backgrounds, catering to individual-level demands and joining the participatory process.

\subsection{Analysis of the collaborative generation}

% \begin{figure}[htbp!]
%     \centering
%     \includegraphics[width=.9\linewidth]{figures/plan_change_hlg.png}
%     \caption{The modification of plan in HLG after revising by each sub-community planner. Change are marked in bold solid lines.}
%     \label{fig:plan_change_hlg}
% \end{figure}

\begin{figure}[htbp!]
    \centering
    \includegraphics[width=.9\linewidth]{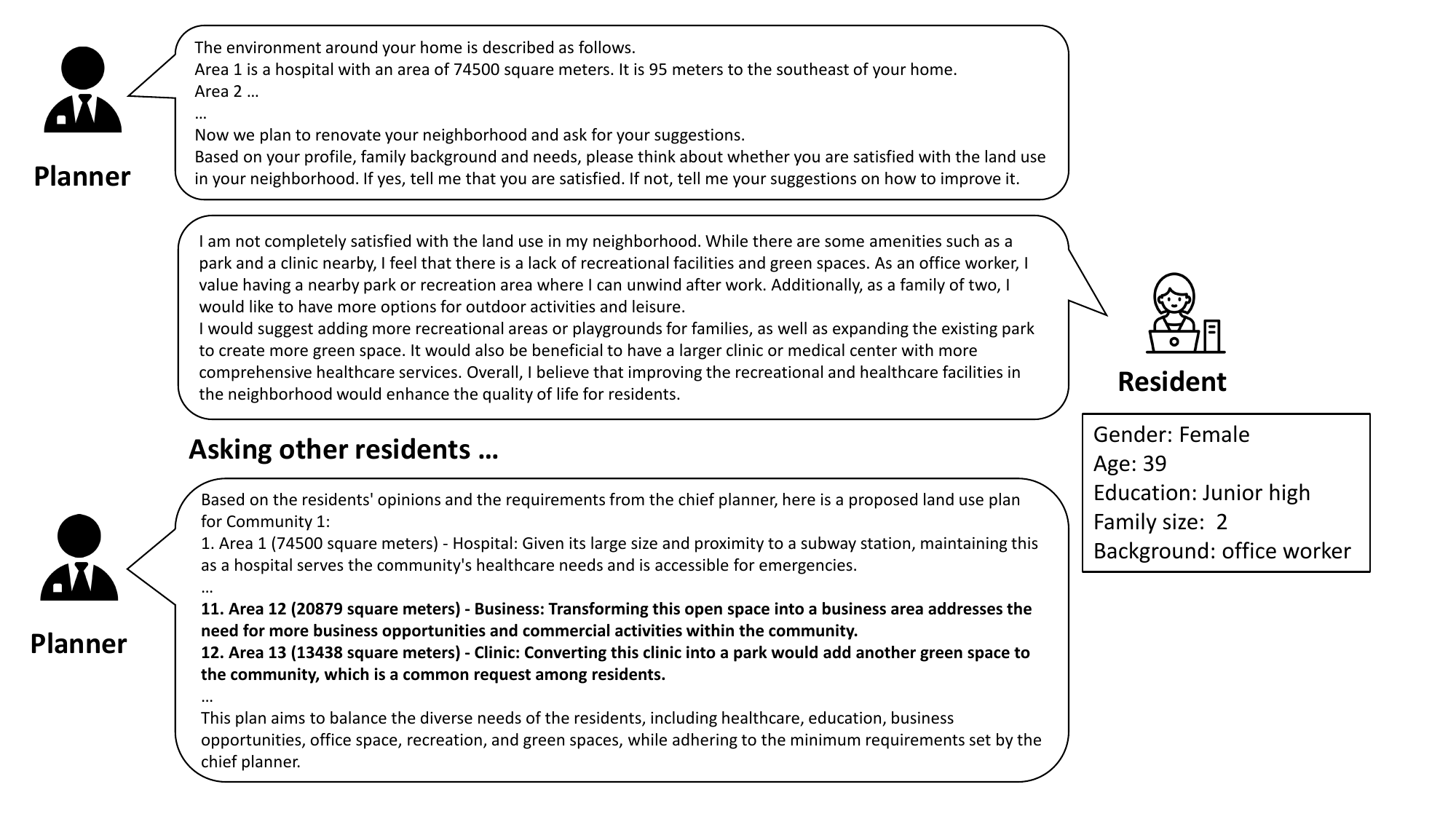}
    \caption{An example dialogue of a sub-community planner discussing with residents and revising the plan.}
    \label{fig:discuss_dialogue}
\end{figure}

Beyond the successful role-play in our framework, we facilitate participatory planning through SPs guiding pertinent resident agents in discussions. Drawing inspiration from the paradigm of focus groups in participatory planning, we partition the entire community into four distinct sub-communities based on the main road network. Resident agents participate in corresponding discussion groups according to the addresses assigned to them. Figure~\ref{fig:discuss_dialogue} illustrates the discussion process, where an SP solicits opinions from each resident and implements corresponding modifications. The aforementioned role-play guides each resident agent to assess their surroundings and provide preferences, \textit{e.g.}, "As an office worker, I value having a nearby park or recreation area where I can unwind after work". In this scenario. SP agents learn from these demands and adjust the planning scheme accordingly.

\begin{figure}[htbp!]
\centering
    \subfigure[HLG]{
    {\label{subfig:plan_change_hlg}}
    \includegraphics[width=.95\linewidth]{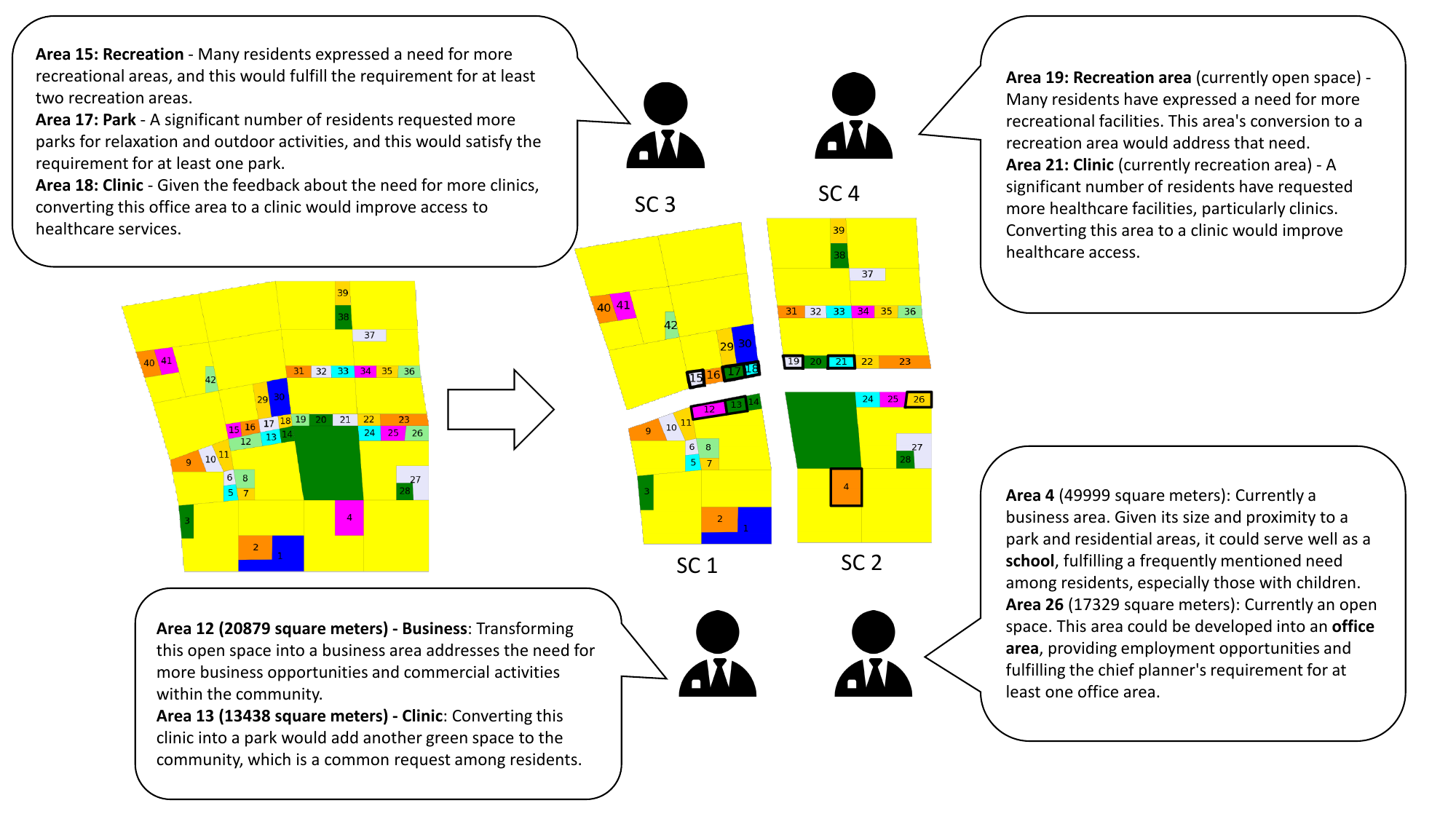}
    }

    \subfigure[DHM]{
    {\label{subfig:plan_change_dhm}}
    \includegraphics[width=.95\linewidth]{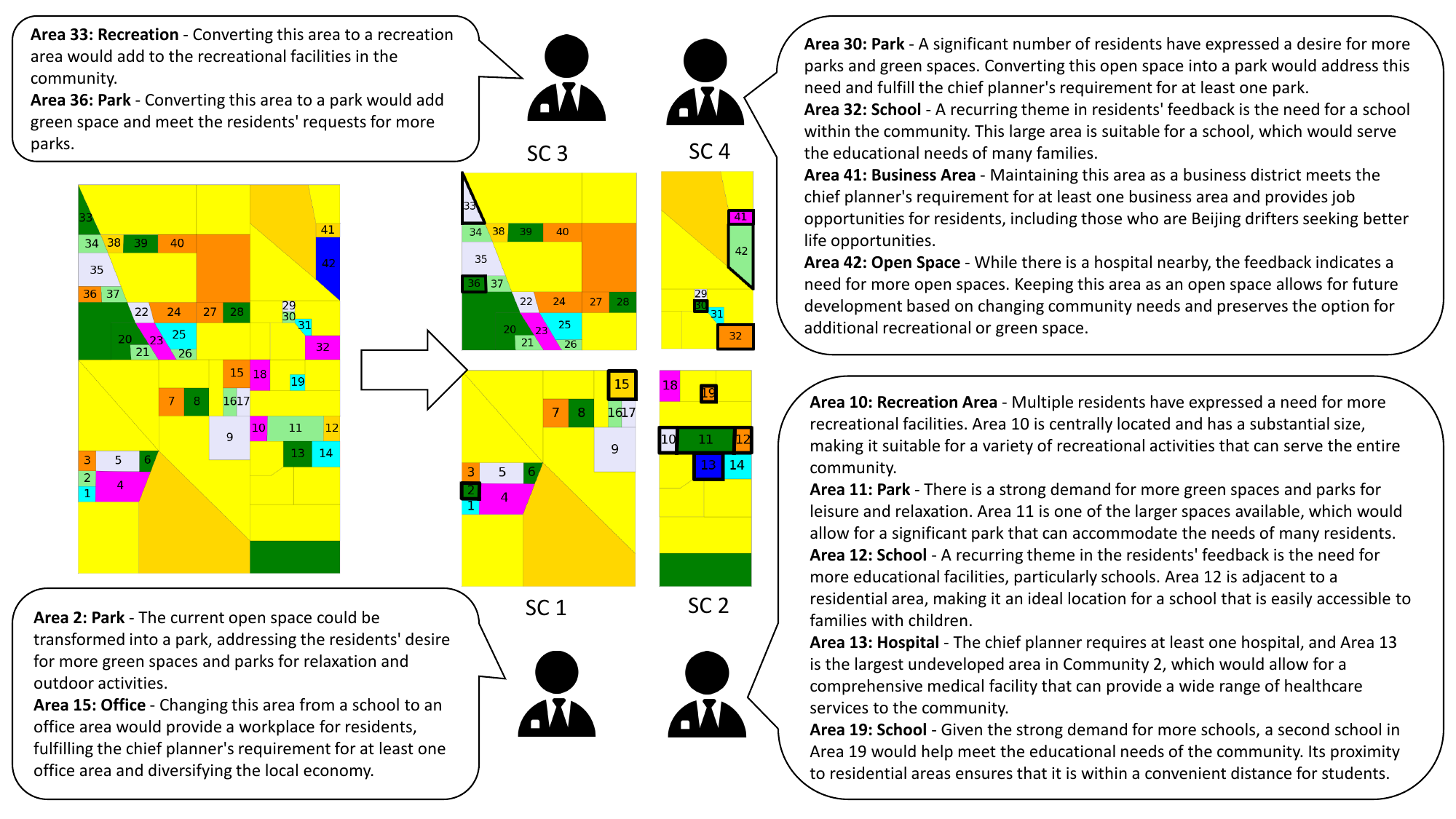}
    }
\caption{The modification of the planning after revising by each sub-community planner and the reasons given by planners. Change are marked in bold solid lines.}
\label{fig:plan_change}
\end{figure}

Figure~\ref{fig:plan_change} shows the substantive change resulting from these discussions, with the initial planning scheme on the left and modifications proposed by each sub-community in bold boxes on the right. Overall, CP generates a reasonably sound initial plan, and participatory discussions refine and enhance this outcome. The hierarchical sub-community structure enables resident agents to concentrate more effectively on areas directly related to their interests and guarantees substantive inclusion of concerns in the planning process.

In the case of HLG (Figure~\ref{subfig:plan_change_dhm}), the revisions predominantly focused on the primary east-west streets. SP agents adjusted land-use types based on residents' needs, such as transforming Area 4 from commercial to educational. This adjustment adheres to planning basic principles, as Area 4 is situated within a residential area with limited commercial value but close to a park. Sub-Community 2 lacks an internal school, justifying the reason given by the SP agent, "Given its size and proximity to a park and residential areas, it could serve well as a school, fulfilling a frequently mentioned need among residents, especially those with children."

Furthermore, SP agents learn to make corresponding adjustments after changes in other sub-communities. For instance, following the transformation of commercial areas into other uses in Sub-Communities 2 and 3 (Area 4 and 15), there is no accessible commercial space in Sub-Community 1. Consequently, Area 12 is transformed from open space to commercial land, "[...] addresses the need for more business opportunities and commercial activities within the community". Similar adjustments can also be observed in DHM (see Figure~\ref{subfig:plan_change_hlg}), where the major hospital relocates across sub-communities. This suggests that our designed discussion mechanism allows feedback across different levels, achieving a globally optimal outcome.

\begin{figure}[htbp!]
\centering
    \subfigure[HLG]{
    {\label{subfig:metric_change_hlg}}
    \includegraphics[width=.45\linewidth]{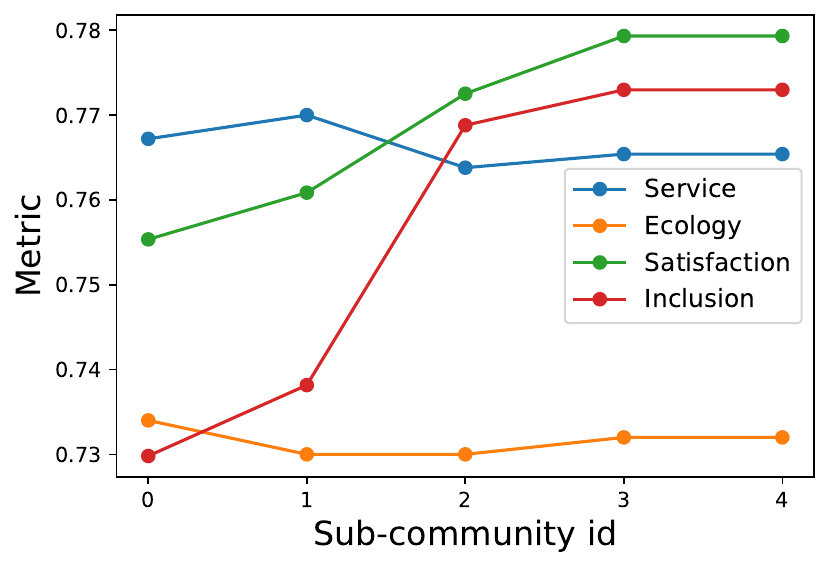}
    }
    \subfigure[DHM]{
    {\label{subfig:metric_change_dhm}}
    \includegraphics[width=.45\linewidth]{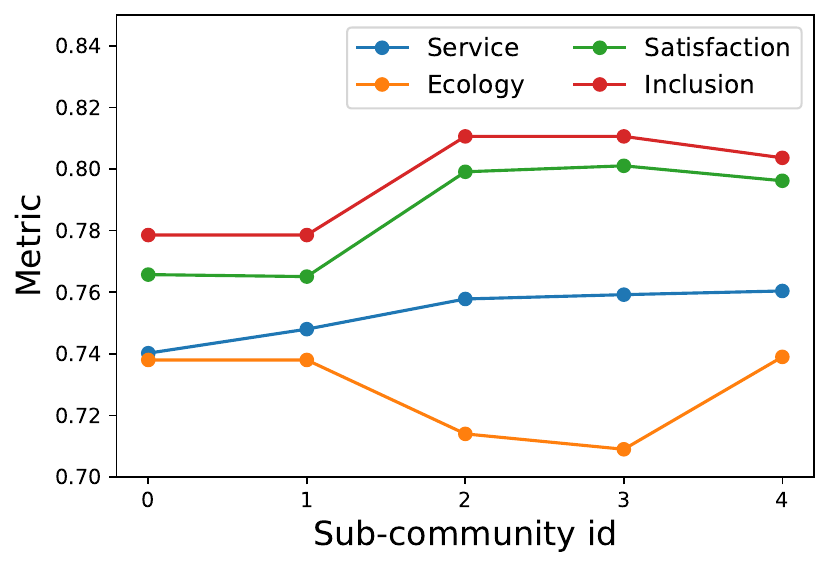}
    }
\caption{The metrics after the revision of each sub-community. 0 indicates the initial plan by the chief planner, and 4 indicates the plan after revising 4 sub-communities, i.e., the final plan.}
\label{fig:metric_change}
\end{figure}

Figure~\ref{fig:metric_change} illustrates the overall improvement in metrics following the successive transformations of the four sub-communities corresponding to HLG and DHM. It can be observed that, with the gradual progression of modifications, need-aware metrics show a noticeable enhancement, while need-agnostic metrics fluctuate. This trend is unsurprising since participatory adjustments primarily consider individual needs. Simultaneously, we observe that these refinements have a minimal impact on need-agnostic metrics, suggesting rational adjustments on the initial planning can significantly improve satisfaction and inclusion without undermining facilities coverage. This underscores the importance of the participatory method.

\subsection{Analysis of the feedback iteration}

\begin{figure}[htbp!]
    \centering
    \includegraphics[width=.9\linewidth]{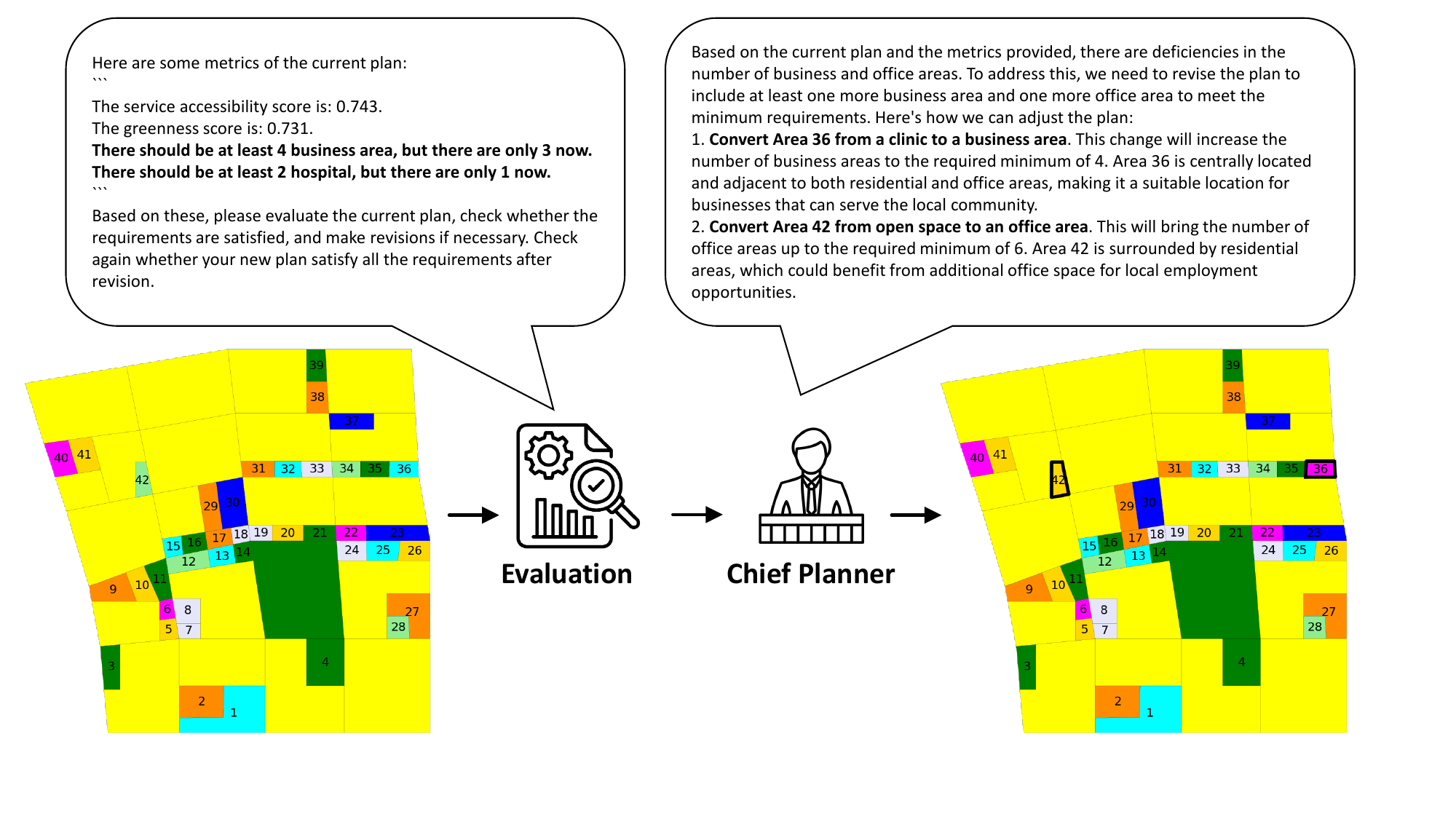}
    \caption{An example of feedback mechanism.}
    \label{fig:feedback}
\end{figure}

As previously analyzed, the plan adjusted by SP may not completely fulfill planning requirements. Consequently, specific modifications are essential to rectify these deficiencies. Figure~\ref{fig:feedback} illustrates that the plan undergoes a fixed mechanism check and evaluation. Violations of requirements are translated into a natural language template, prompting CP to make corresponding adjustments. The CP agent will address conflicting portions, considering the overall plan and the characteristics of each plot, as highlighted by black boxes.

The revision reveals that the CP agent possesses a nuanced understanding of the community's characteristics. It underscores, for instance, that "Area 42 [...] could benefit from additional office space for local employment opportunities," aligning with the context of HLG being a high-density area with insufficient job opportunities.
\section{Discussion}

% 限制：
% Agent代替真人反映诉求的有效性
% 上位规划的限制、规划原则的限制

In conclusion, our research introduces a novel urban planning approach that integrates role-playing, collaborative generation, and feedback iteration using LLM agents in the participatory paradigm. Through empirical experiments in two distinct communities HLG and DHM, we demonstrate that our LLM-powered model performs exceptionally well, ranking second only to RL in service (0.756 and 0.760) and ecology (0.714 and 0.739) metrics, while achieving optimal results in satisfaction (0.784 and 0.784) and inclusion (0.764 and 0.794). Our findings highlight the capacity of LLM to generate coherent urban planning schemes, with a detailed analysis emphasizing its ability to address individual needs, incorporate opinions substantively, and deliver outcomes that are transparent and adaptable. The main contributions made are listed below:

\begin{itemize}
    \item \textbf{Innovative LLM-powered participatory framework:} We propose a novel participatory framework that leverages the capacity of LLM, showcasing the potential of LLMs in generating coherent and participatory planning schemes.
    \item \textbf{Adaptability in diverse urban contexts:} We conduct empirical experiments in two diverse urban communities, demonstrating the adaptability and effectiveness of the LLM-powered model across varied urban planning scenarios.
    \item \textbf{Compelling performance outcomes:} The research yields compelling performance outcomes, showcasing that the LLM-powered model achieves more competitive results than human experts and is comparable to the newest RL method. 
\end{itemize}

It is also noteworthy that, as a newly emerging generative model, planners and residents can benefit from LLM agents. For planners, the resident agents played by LLM, though not necessarily mirroring the real world, sufficiently reflect potential needs and possible challenges. Especially this form of role-playing, devoid of time-consuming iterative discussions, offers a convenient means for participatory discussion and presents a simple emulation from tokenism consultative steps to citizen-controlled collaboration~\cite{arnstein1969ladder}. For residents, the planner agent played by LLM theoretically possesses vast planning knowledge from experts, as we have seen in the online health consultation~\cite{howard2023chatgpt}. Furthermore, the planner agent can transparently and interactively elucidate all motivations and reasons behind the planning decisions in natural language, simultaneously eliminating potential biases and conflicts of interest. Therefore, LLMs-empowered agents form a low-cost, highly efficient approach crucial in enhancing participation in planning and rebuilding trust between residents and professionals.

We must acknowledge the numerous simplifications regarding planning elements involved throughout the process. Factors such as ownership, development costs, higher-level planning, or community visions were omitted in this study, potentially leading to a deviation from practical planning scenarios. However, the immense scalability of LLM based on natural language allows these shortcomings to be addressed through new prompts without compromising the effectiveness and interpretability of our framework. Similarly, the extent to which LLM agents can authentically emulate humans remains a question awaiting further advancements in computer technology. We must also note that our objective is not to replace real humans with LLM agents, and we acknowledge citizen engagement is paramount within the participatory planning framework. Our anticipation is that LLM agents, serving as planning support systems, can generate potential opinions, provide professional knowledge, assist in citizen participation, ultimately foster trust between citizens and professionals, and facilitate the realization of participatory urban planning.

%% The Appendices part is started with the command \appendix;
%% appendix sections are then done as normal sections
%% \appendix

%% \section{}
%% \label{}

%% If you have bibdatabase file and want bibtex to generate the
%% bibitems, please use
%%
\bibliographystyle{elsarticle-num} 
\bibliography{reference}

\end{document}